\documentclass[10pt,twocolumn,letterpaper]{article}

\usepackage[pagenumbers]{cvpr} 

\usepackage{float}
\usepackage[dvipsnames, table]{xcolor}
\usepackage{duckuments}
\usepackage{bm}

\definecolor{first}{rgb}{1.0, .83, 0.3}
\definecolor{second}{rgb}{1.0, 0.93, 0.7}
\def \first {\cellcolor{first}\bfseries}
\def \second {\cellcolor{second}}

\definecolor{cvprblue}{rgb}{0.21,0.49,0.74}
\usepackage[pagebackref,breaklinks,colorlinks,citecolor=cvprblue]{hyperref}

\usepackage{float}
\usepackage{tabularx}
\usepackage{multirow}
\usepackage{epigraph}
\usepackage{amsmath}
\usepackage{amssymb}
\usepackage{cuted}
\usepackage{url}
\newcommand{\ours}[0]{Turbo3D}

\def\paperID{5775} 
\def\confName{CVPR}
\def\confYear{2025}

\title{Turbo3D: Ultra-fast Text-to-3D Generation}

\author{
Hanzhe Hu$^{1}$\and
Tianwei Yin$^{2}$\and
Fujun Luan$^{3}$\and
Yiwei Hu$^{3}$\and
Hao Tan$^{3}$\and
Zexiang Xu$^{3}$\and
Sai Bi$^{3}$\and
Shubham Tulsiani$^{1}$\thanks{Equal advising.}\and 
Kai Zhang$^{3}$\footnotemark[1] \and
\\
$^{1}$ Carnegie Mellon University \quad
$^{2}$ Massachusetts Institute of Technology \quad
$^{3}$ Adobe Research \quad\\
\\ {\tt \small \href{https://turbo-3d.github.io/}{https://turbo-3d.github.io/}}
}

\begin{document}
\maketitle

\begin{strip}
\vspace{-4em}
    \centering
    \includegraphics[width=1.\linewidth]{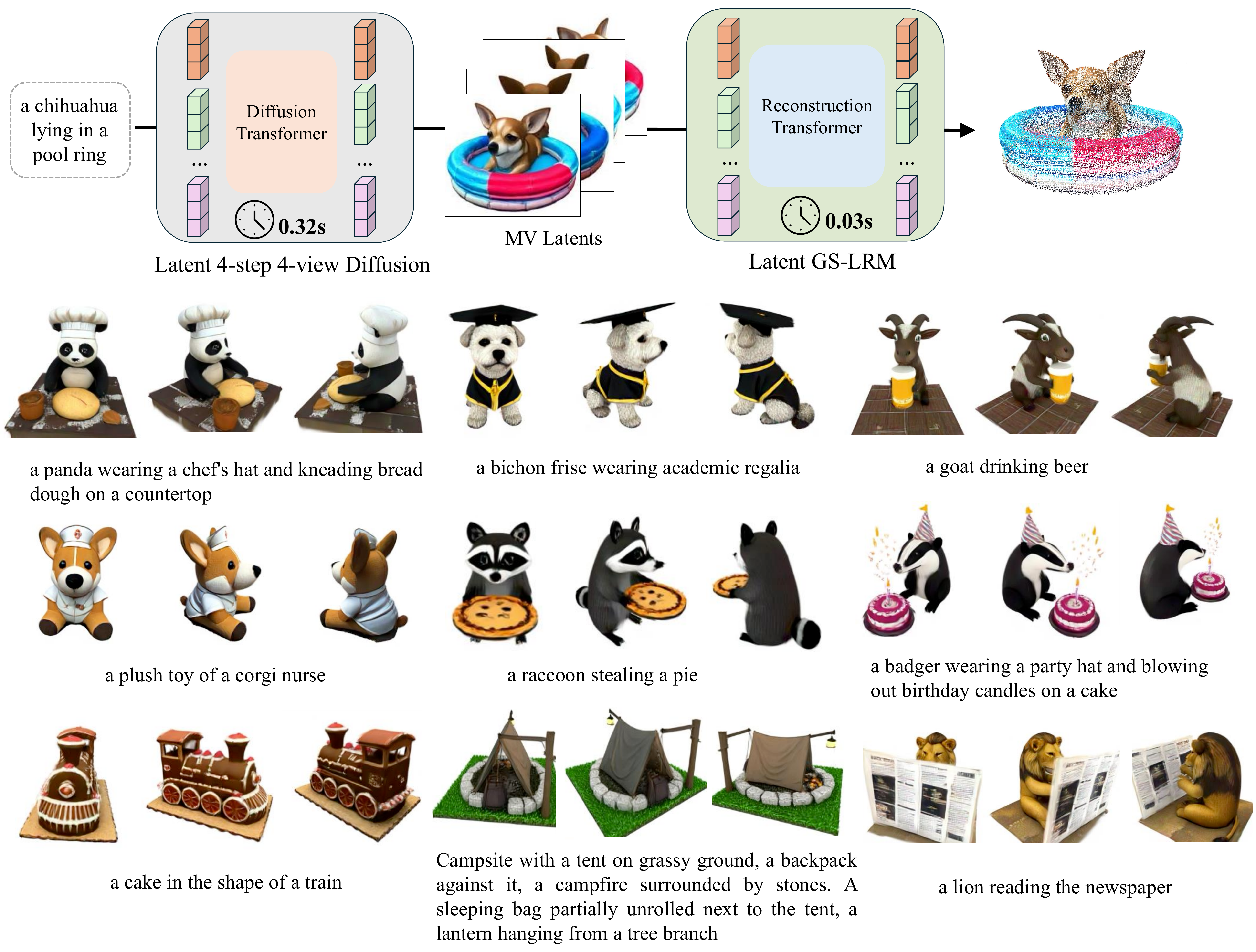}
    \captionof{figure}{\textbf{Overview of our \ours\ text-to-3D system. } \ours~  generates high-quality 3D Gaussian Splatting (3DGS) assets from user prompts in \textbf{\textit{less than 1 second}} on a single A100 GPU. It's a two-stage pipeline consisting of a highly efficient latent-space few-step multi-view (MV) generator and single-step MV reconstructor. Note that we visualize latents as RGB images and 3DGS assets as point clouds   in the pipeline figure for clarity. }
    \label{fig:teaser}
\end{strip}

\begin{abstract}
We present \ours, an ultra-fast text-to-3D system capable of generating high-quality Gaussian splatting assets in under one second. \ours\ employs a rapid 4-step, 4-view diffusion generator and an efficient feed-forward Gaussian reconstructor, both operating in latent space, as shown in Fig.~\ref{fig:teaser}. The 4-step, 4-view generator is a student model distilled through a novel Dual-Teacher approach, which encourages the student to learn view consistency from a multi-view teacher and photo-realism from a single-view teacher. By shifting the Gaussian reconstructor's inputs from pixel space to latent space, we eliminate the extra image decoding time and halve the transformer sequence length for maximum efficiency.
Our method demonstrates superior 3D generation results compared to previous baselines, while operating in a fraction of their runtime. 

\end{abstract}

%
\section{Introduction}
\label{sec:intro}


The recent advances in image generative models allow users to generate detailed outputs from just a text prompt. 
While initial denoising diffusion-based methods~\cite{vincent2011connection, ho2020denoising, song2020score, rombach2022high, ramesh2021zero, saharia2022photorealistic} enabled impressive photo-realistic generation, recent techniques~\cite{yin2024one, yin2024improved, liu2022flow, lipman2022flow, liu2023instaflow, song2023consistency, sauer2024fast} have significantly improved inference efficiency of such models, allowing high-fidelity generation in the blink of an eye. Unfortunately, these advances in generative 2D modeling methods have not yet been matched in the 3D domain,~where ultra-fast realistic 3D generation remains a challenge. In this work, we seek to bridge this gap, and present \ours, a text-to-3D generative model that can synthesize detailed 3D outputs in a fraction of a second.


The existing approaches for text-based 3D inference can be categorized as either generative~\cite{jun2023shap, nichol2022point,  li2023instant3d, liu2023syncdreamer, zhang2024clay, wu2024direct3d} or optimization-driven~\cite{poole2022dreamfusion, wang2024prolificdreamer, lin2023magic3d}. The latter class of methods optimize 3D representations by `distilling' pre-trained 2D diffusion models~\cite{poole2022dreamfusion}. 
While this approach can yield decent 3D outputs, it
is highly inefficient, typically requiring several minutes or even hours to output a single 3D representation. 
The alternative paradigm is to learn a generative model that directly outputs 3D representations. While initial methods~\cite{jun2023shap,autosdf2022} investigated representations such as point clouds and SDFs, the limited availability of 3D data restricted the generation quality. Recent approaches~\cite{liu2023syncdreamer, shi2023zero123++, shi2023mvdream, li2023instant3d} have instead advocated for learning generative models of multi-view images (followed by deterministic 3D reconstruction), as these can be initialized from pre-trained 2D generative models. While these methods have resulted in impressive generations, the multi-view finetuning on synthetic data does inhibit their generation quality. More crucially, their inference efficiency is restricted by the iterative denoising process required for the text-conditioned multi-view generation.


In this work, we follow the paradigm of text-to-3D via multi-view generation, and aim to improve the efficiency of the underlying components to enable ultra-fast generation, while enhancing the fidelity of the generated outputs.
Inspired by the recent progress in reducing inference time by distilling 2D diffusion models into one-step or few-step generators~\cite{salimans2022progressive,song2023consistency,yin2024one,sauer2024fast}, we  adapt these for multi-view 3D generation. We first train a many-step text-to-multi-view diffusion model and then distill it into a much faster 4-step generator using the distribution matching distillation (DMD) loss~\cite{yin2024one}. 
However, we find that this process leads to a significant quality degradation, as it fails to capture the full range of modes present in the multi-view teacher model. 
To overcome this, we propose to extend the DMD pipeline to incorporate another single-view teacher -- a 2D denoising diffusion model trained on a large set of high quality aesthetic images. Our few-step multi-view generator is thus trained with a dual-teacher distillation approach, where the multi-view DMD loss helps our model learn multi-view consistency,
and the single-view DMD loss ensures high-fidelity outputs. To further improve the 3D generation efficiency, we note that our multi-view generator outputs latent representations (and not pixels) for the multi-view images. We build on this insight to adapt a prior multi-view to 3D reconstruction approach~\cite{zhang2025gs} to instead consume multi-view latents as input, and show that this improves the reconstruction efficiency without any performance loss. 


Our overall system thus combines an efficient few-step multi-view generator with a multi-view latent to 3D model to enable ultra fast 3D generation. We train our model on the subset of Objaverse dataset~\cite{deitke2023objaverse, deitke2024objaverse}, which contains about 400k instances with Cap3D text captions~\cite{luo2024scalable}. We show that our proposed \ours\ is able to produce high-quality 3D assets in less than one second (see Fig.~\ref{fig:teaser}), and also achieves comparable quality with previous state-of-the-art.


\section{Related Work}
\label{related}

In this section, we discuss the closely related prior feed-forward 3D generation methods. Optimization-based methods~\cite{poole2022dreamfusion, lin2023magic3d} is out of this work's scope. 
We review feed-forward 3D generative models in the following three categories: 1) methods that directly generate full 3D representation encoding geometry and appearance; 2) methods that generate shapes and then generate the textures; 3) methods that generate multi-view (MV) images followed by reconstruction.  
We also review the diffusion distillation literature which our method is built upon.



\subsection{Directly Genereating Full 3D Representation} 
Several prior methods~\cite{jun2023shap, nichol2022point, wang2023rodin, anciukevivcius2023renderdiffusion, xu2023dmv3d, szymanowicz2023viewset, tewari2023diffusion} have been introduced that directly generate 3D representations encoding both geometry and texture, e.g., implicit fields~\cite{jun2023shap}, point clouds~\cite{nichol2022point}, and triplanes~\cite{wang2023rodin}. This line of work typically have to preprocess the source 3D data (meshes or multi-view images) into the target representations for generative models in a lossy fashion, hence limiting their quality and scalabilty. 

Alternative methods~\cite{anciukevivcius2023renderdiffusion, xu2023dmv3d, szymanowicz2023viewset, tewari2023diffusion} have been proposed to get rid of the lossy preprocessing. They implicitly bake a 3D generative process into a multi-view diffusion framework, creating a single-stage 3D generative model. However, trained only on 3D data, they suffer from the issue of limited generalization, lower pixel quality and reduced understanding of complex text prompts, compared with methods leveraging powerful pretrained text-to-image models~\cite{liu2023zero, li2023instant3d}.  

Our method leverages the strong priors in a pretrained image generative model, but we go a step further by distilling the slow teacher model into a fast student generator for ultra-fast text-to-3D. 

 


\subsection{Shape Generation + Texture Generation}
Recent advancements in 3D content creation have introduced innovative methods~\cite{zhang2024clay, wu2024direct3d, siddiqui2024meshgpt} that focus on generating high-quality 3D shapes first, followed by generative texturing~\cite{richardson2023texture, chen2023text2tex}, rather than creating both simultaneously. CLAY~\cite{zhang2024clay} offers a framework using a multi-resolution Variational Autoencoder (VAE) and a latent Diffusion Transformer (DiT) to initially create detailed 3D geometries from inputs like text and images, before applying high-resolution physically-based rendering (PBR) textures. Direct3D~\cite{wu2024direct3d} enhances scalability in image-to-3D generation by improving the performance of 3D VAE and DiT for generating 3D shapes, subsequently adding textures.

However, besides the challenge of learning a compact latent space suitable for generation, these methods are also usually slow as a result of the two diffusion models - one for shapes and the other for textures. In contrast, our \ours\ only have one diffusion model for generating multi-views and is fast once distilled, while the reconstructor is deterministic and hence fast.


\subsection{MV Generation + MV Reconstruction}
A large body of the 3D generation work~\cite{li2023instant3d,shi2023mvdream, liu2023syncdreamer, liu2024one} focus on leveraging multi-view generation and reconstruction to enhance quality and efficiency.  Instant3D~\cite{li2023instant3d} offers a fast method for creating high-quality 3D assets from text prompts by combining sparse-view generation with a transformer-based reconstructor~\cite{hong2023lrm,wang2023pf,zhang2025gs,wei2024meshlrm}, significantly reducing inference time. MVDream~\cite{shi2023mvdream} employs a multi-view diffusion model to improve consistency and stability in 3D generation, integrating 2D and 3D data. SyncDreamer~\cite{liu2023syncdreamer} enhances multiview-consistency from a single-view image using a 3D-aware feature attention mechanism. One-2-3-45~\cite{liu2024one} introduces an efficient approach to single image 3D reconstruction without extensive optimization, producing consistent 3D meshes. SV3D~\cite{voleti2024sv3d} utilizes a latent video diffusion model for novel multi-view synthesis, incorporating explicit camera control to improve 3D reconstruction quality.

We follow the same approach of reconstructing generated multi-views to create 3D assets in this work, but we focus on improving the efficiency of such systems. With our novel Dual-Teacher Distillation and Latent GS-LRM components, our \ours\ manages to be at least an order of magnitude faster than these baselines while maintaining competitive quality. Concurrent work  GECO~\cite{wang2024geco} also use diffusion distillation to speed up image-to-3D. However, we  differ in our dual-teacher distillation design with focus on text-to-3D. Our pipeline is also simpler and avoids the cumbersome mesh reconstructions for 3D distillation in GECO. 



\subsection{Diffusion Distillation}
Recent progress in diffusion models have focused on improving the efficiency of image generation by reducing the number of sampling steps required, leading to the development of several innovative distillation techniques~\cite{yin2024improved,yin2024one,song2023consistency,salimans2022progressive,xu2024ufogen,nguyen2024swiftbrush,luo2024diff,sauer2024fast,meng2023distillation}. 
Methods like Improved Rectified Flows~\cite{lee2024improving} and InstaFlow~\cite{liu2023instaflow} straighten the ODE trajectories, making them easier to approximate with a one-step student model. Consistency Models~\cite{song2023consistency} train student generators to map any point on the teacher's ODE trajectory to a consistent target, enabling one-step and few-step image generation. Distribution Matching Distillation (DMD)~\cite{yin2024one} trains a one-step generator by minimizing the reverse KL divergence between the data distribution and the generator's output distribution~\cite{wang2024prolificdreamer, luo2024diff, franceschi2024unifying}. 
Building on this, DMD2~\cite{yin2024improved} integrates GAN losses~\cite{goodfellow2014generative} and extends the method to multi-step generators, further enhancing generation quality.

However, most of existing approaches focus on distilling pretrained 2D image diffusion models. In our work, we adopt the popular DMD~\cite{yin2024one} approach and extend it to the multi-view domain. Distilling multi-view models presents unique challenges, such as increased mode collapse due to fine-tuning and distillation processes. To address this issue, we introduce a novel dual-teacher distillation technique, enabling swift, photorealistic multi-view generation.

\section{Background}

\subsection{Multi-view Diffusion Model}
Diffusion models generate data from noise by reversing a forward noising process. 
Given a data sample $x_0 \sim p(x_0)$ from the data distribution, the forward process progressively adds Gaussian noise over $T$ timesteps. 
At timestep t, the noised distribution conditioned on $x_0$ is given by:
\begin{equation} q(x_t | x_0) = \mathcal{N}(x_t; \alpha_t x_0, \beta_t^2 \mathbf{I}), \label{eq:forward_diffusion}
\end{equation}
where $\alpha_t$ and $\beta_t$ are timestep-dependent constant that are specified by the noise schedule~\cite{song2020score, kingma2021variational}. 
The diffusion model $\epsilon_\theta$ is trained to reverse this noising process by learning to predict the added Gaussian noise $\epsilon$:
:
\begin{equation} L(\theta) = \mathbb{E}  || \epsilon - \epsilon_\theta\left(x_t, t \right) ||^2. 
\end{equation}
There exists other formulations, in which the diffusion model learns to predict the clean input $x_0$ directly~\cite{karras2022elucidating}, or a combination of $x_0$ and $\epsilon$~\cite{salimans2022progressive}.
Regardless of the specific prediction target, the outputs of these models can be related to the score function, which is the gradient of the log probability density of the data distribution~\cite{song2020score}: 
\begin{equation}
\small 
    s_\theta(x_t, t) = \nabla_{x_t} \text{log} \ p(x_t) = -\frac{\epsilon_\theta(x_t, t)}{\beta_t} = -\frac{x_t-\alpha_t x_\theta(x_t, t)}{\beta_t^2} 
\end{equation}
where $\epsilon_\theta$~\cite{ho2020denoising} and $x_\theta$~\cite{karras2022elucidating} represents noise and data prediction diffusion models, respectively. 
In our paper, we adopt the noise prediction scheme but our method generalizes to any formulations. 

In multi-view diffusion models, the generation of a 3D scene is conditioned on a text prompt, enabling the joint denoising of multiple views to produce a set of 3D-consistent output images~\cite{li2023instant3d, shi2023mvdream}. Specifically, given a text prompt c and a set of multi-view images $\{x^i\}_{i=1}^K$ where $K$ is the number of view, the multi-view diffusion model learns to predict the added noise across all views simultaneously.
The training objective is formulated as:
\begin{equation}
\mathbb{E}  || \epsilon - \epsilon_\theta(\{x^i_t\}_{i=1}^K, t, c)||^2.  
\label{eq:denoising_loss}
\end{equation}
where $\epsilon$ represents independent Gaussian noise with the same variance applied to each view. 
At inference time, the generation starts from a set of fully noisy multi-view images $\{x_T^i\}_{i=1}^K$ sampled from a standard Gaussian distribution. 
The multi-view diffusion model iteratively generate a sequence of cleaner multi-view images.
Various diffusion samplers~\cite{ho2020denoising, song2020denoising} can be employed during this process to optimize generation speed and quality.

\subsection{Distribution Matching Distillation}
Distribution Matching Distillation (DMD) is a widely used diffusion distillation technique that converts teacher diffusion models into a student generator requiring significantly fewer sampling steps~\cite{yin2024improved, yin2024one}. The DMD approach trains the student generator $G_\theta$ by minimizing an approximate reverse KL divergence between the smoothed student's output distribution (denoted as $p_\text{fake}$) and the smoothed data distribution (denoted as $p_\text{real}$):
\begin{equation}
\small 
L_{\text{DMD}}(\theta) = D_{\text{KL}}(p_\text{fake} || p_\text{real}) = \mathbb{E}_{x,t} \ ( \text{log}(\frac{p_\text{fake}(x_t)}{p_\text{real}(x_t)}) ). \end{equation}
Let the score functions of the data distribution and the student's output distribution be denoted as $s_\text{real}$ and $s_\text{fake}$, respectively.  
The gradient of this KL divergence can be effectively approximated by the difference between these two score functions:
\begin{equation}
\begin{split}
    \nabla_\theta L_{\text{DMD}}(\theta) \approx \mathbb{E} \Bigg[ \ 
    & -\int \left( s_\text{real}\big(F(G_\theta(\epsilon), t), t\big) \right. \\
    & \left. \quad - \, s_\text{fake}\big(F(G_\theta(\epsilon), t), t\big) \right) \frac{dG_\theta(\epsilon)}{d\theta} d\epsilon   
    \Bigg]
\end{split}
\end{equation}
where $F$ represents the forward diffusion process defined in Eq.~\ref{eq:forward_diffusion}. 
The student generator can be adapted for a multi-step generation setting by replacing the pure noise input $\epsilon$ with a partially noisy image $x_t$~\cite{yin2024improved}. 
During training, the data distribution's score function is initialized from the teacher diffusion model and kept fixed, while the student's output distribution score function is dynamically trained using the student’s output and a denoising loss~(Eq.~\ref{eq:denoising_loss}).


\section{Method}

\begin{figure*}[ht]
    \centering
    \includegraphics[width=1.\linewidth]{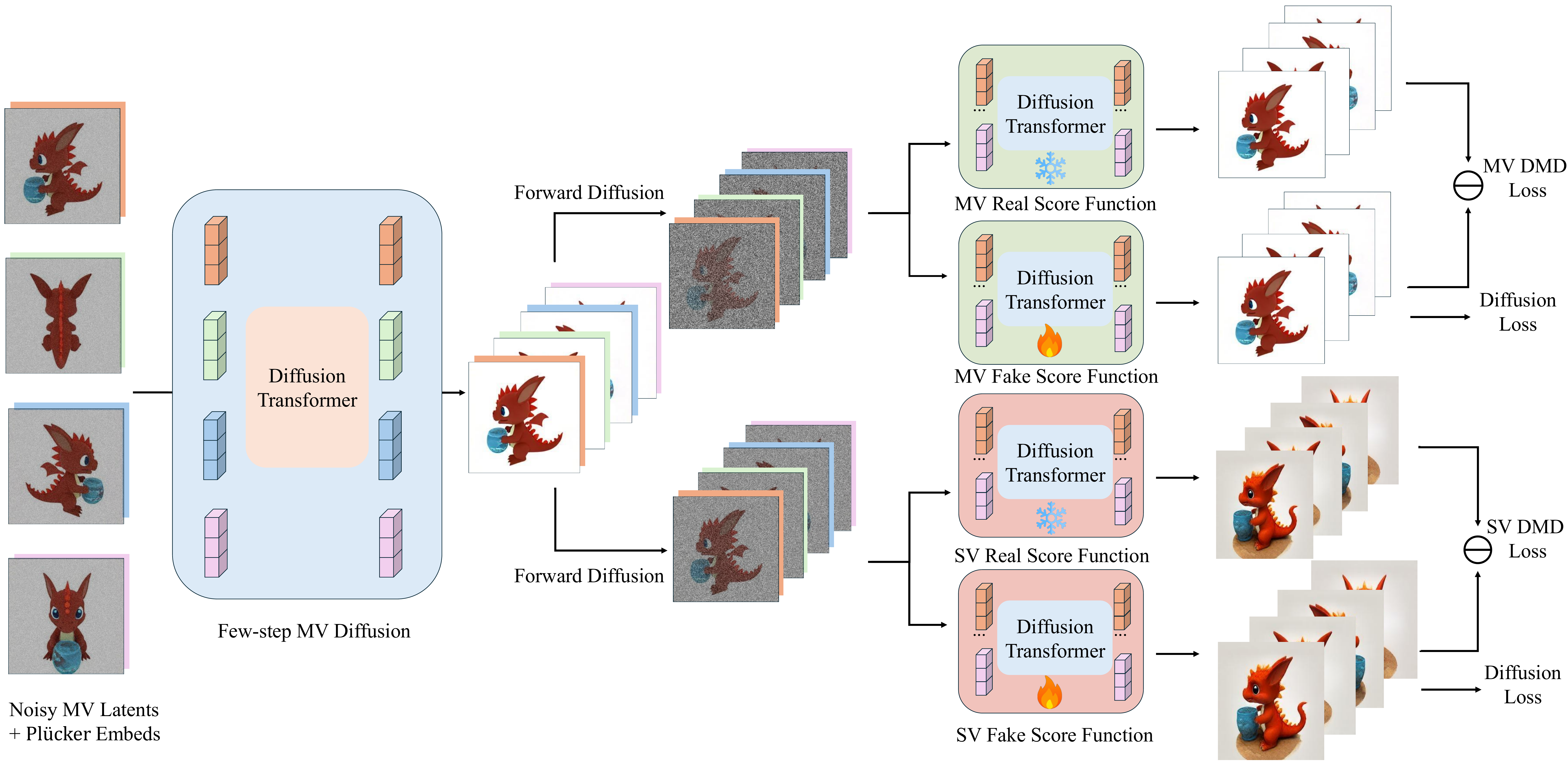}
    \caption{\textbf{Dual-teacher distillation framework in our \ours.} Note that latents are visualized as RGB images for clarity. We aim to distill a multi-step multi-view teacher generator (right, green) into a few-step multi-view generator (left, blue). Our few-step MV student generator is conditioned on Plücker embeddings for better 3D awareness. Similar to ~\cite{yin2024improved}, we optimize the student generator using distribution matching objective (DMD loss) and train the fake score function to model the distribution of samples produced by the student generator. In particular, we integrate two teacher models: multi-view teacher and single-view (SV) teacher to enhance both multi-view consistency and photorealism. The MV score functions take a set of images of one object as input and calculate the MV DMD loss, while the SV score functions treat each image separately and calculate the SV DMD loss. }
    \label{fig:dualteacher}
\end{figure*}


\label{sec: method}
In this section, we provide the technical details of our \ours\ text-to-3D system, which features a highly efficient multi-view (MV) generator and reconstructor. We begin by describing our novel DualTeacher Distillation approach; it creates a rapid MV generator by jointly distilling knowledge from both a multi-view (MV) teacher and a single-view (SV) teacher. Following this, we discuss the latent-space GS-LRM that instantly lifts the generated MV latents into high-quality 3D Gaussians.




\subsection{Dual-teacher Distillation for MV Diffusion} \label{sec:dual-teacher}
Leveraging a multi-step multi-view diffusion in a text-to-3D generation process can be inefficient due to repeated evaluations of the diffusion denoiser in the sampling process. To speed this up, one approach is to use diffusion distillation~\cite{yin2024one, yin2024improved} to train a single-step or few-step MV generator.


However, we find that naively applying diffusion distillation methods to a MV teacher can cause the student model to generate overly simplistic and cartoonish appearance, which closely resembles the 3D Objaverse dataset used during MV teacher finetuning and distillation~\cite{deitke2023objaverse} (shown in Fig.~\ref{fig:ablation}). We call this phenomenon `\textbf{compounding mode collapse}' 
; this happens because both finetuning and distillation sacrifices generation diversity for efficiency. As the MV teacher is already biased towards synthetic Objverse-style appearance, further distilling it will have compound effect of locking the distilled generator in the mode of Objaverse data that are far from the modes of photorealistic natural images.


To address this issue, we propose to use dual teachers in the distillation process: one MV teacher to teach the student model about multi-view consistency, and one SV teacher to teach about each views' photo-realism. We illustrate this Dual-teacher Distillation algorithm in Fig.~\ref{fig:dualteacher}. Concretely, this is formulated as:

\begin{align}
\small 
L_{\text{DMD}}^{\text{Dual}}(\theta) = &D_{\text{KL}}\big(p_\text{fake}(\{x_t^i\}_{i=1}^{K})\ ||\ p_\text{real}^{\text{MV}}(\{x_t^i\}_{i=1}^{K})\big) \nonumber \\
&+ \lambda \cdot \frac{1}{K}\sum_{i=1}^{K} D_{\text{KL}}\big(p_\text{fake}(x_t^i)\ ||\  p_\text{real}^{\text{SV}}(x_t^i)\big),
\end{align}
where $p_\text{real}^{\text{MV}}(x_t), p_\text{real}^{\text{SV}}(x_t)$ represent MV and SV teachers, respectively; $\lambda$ is the loss weight balancing the influence of MV and SV teachers on the distilled student model; $K=4$ is the number of views.  We set $\lambda$ to 1 in our experiments. As demonstrated in Fig.~\ref{fig:ablation}, having the additional SV teacher in the distillation process effectively addresses the compound mode collapse issue, because it tries to pull each view to look like natural images. 


\subsection{Latent GS-LRM for MV Reconstruction}
To reconstruct 3D from the generated MV latents, one straightforward approach is to decode them into multi-view images, and then use the pixel-space GS-LRM~\cite{zhang2025gs} to produce the 3D Gaussians. However, such an approach can suffer from efficiency and memory issue when scaling to high resolution due to the poor performance of Conv2D operators in VAE decoder~\cite{hanlab_patchconv_2024}.

\begin{figure}
     \centering
     \includegraphics[width=1.0\linewidth]{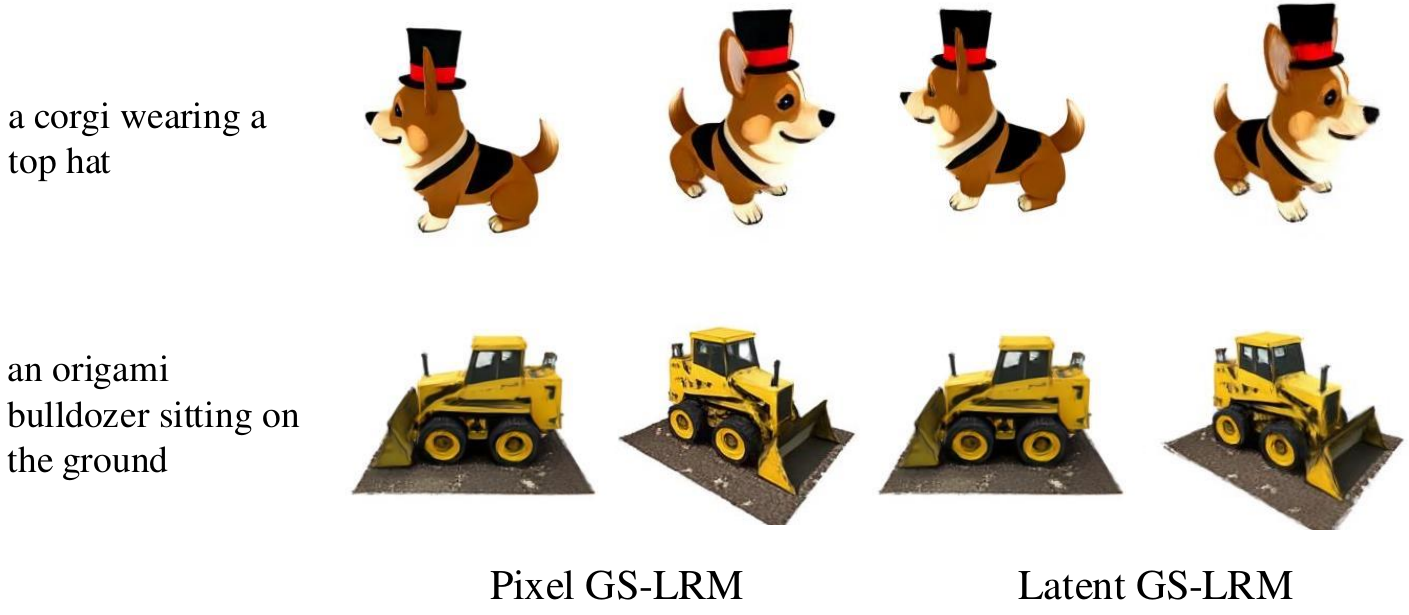}
     \caption{We compare the renderings of pixel GS-LRM and latent GS-LRM. Latent GS-LRM achieves comparable reconstruction quality as pixel GS-LRM.}
     \label{gslrm_comp}
 \end{figure}

We propose to skip the VAE decoding and directly input the generated MV latents to a latent GS-LRM for best efficiency. To train such a model, 
 we supervise the reconstructed Gaussians with pixel-space novel-view rendering losses ($\ell_2$ and perceptual losses as in \cite{zhang2025gs}). We show that replacing pixel-space GS-LRM with a latent one does not affect the quality of generated assets in Tab.~\ref{abla-gslrm} and Fig.~\ref{gslrm_comp}
 , while being faster. 

\section{Experiments}

In this section, we first describe our experimental setup (Section~\ref{setup}). We then compare our method with state-of-the-art text-to-3D baselines (Section~\ref{main_res}). Finally, we ablate each component of our framework to showcase their effectiveness (Section~\ref{ablation-study}).

\subsection{Experimental Setup}
\label{setup}

\noindent\textbf{Datasets. }
We use the Objaverse dataset~\cite{deitke2023objaverse} to train both our multi-view generation model and multi-view reconstruction model. We scale the objects and center them to fit into a cube $[-1,1]^3$. For the generation task, we render the dataset at a fixed elevation (20 degrees) and 16 equidistant azimuths to achieve a good coverage of the object. We render using a field of view $50^o$ at a distance of 2.7 and uniform lighting. For the reconstruction task, we render 32 views randomly placed around the object with a random distance in the range of $[1.5, 2.8]$. We render a total of $730K$ objects.

\noindent\textbf{Baselines. }
We adopt Instant3D~\cite{li2023instant3d} and LGM~\cite{tang2025lgm} as our baseline text-to-3D   methods. However, since the field of fast text-to-3D is relatively underexplored, we also include recent state-of-the-art fast image-to-3D methods TripoSR~\cite{tochilkin2024triposr} and SV3D~\cite{voleti2024sv3d} as baseline methods. For a fair comparison, we use a popular few-step text-to-image model Flux~\cite{blackforestlabs_flux_2024} to generate an input image first and then feed it to the image-to-3D models. The inference time of image-to-3D models is a summation of the two parts.

\noindent\textbf{Metrics. }We adopt the CLIP score~\cite{radford2021learning} and VQA score~\cite{li2024naturalbench} to assess the semantic alignment between the generated results and text prompts. We use 400 text prompts from DreamFusion~\cite{poole2022dreamfusion} for evaluation. We generate one object for each prompt. For each generated 3D object, we render 10 random views and calculate the average CLIP score and VQA score between the rendered images and the input text. For inference time, we report it with all methods under the same image resolution of 256. Notably, some methods, e.g., Instant3D~\cite{li2023instant3d}, only support a higher resolution of 512. For a fair comparison, we report their quantitative results under the resolution of 512 and only inference time on the resolution of 256.

\noindent\textbf{Implementation Details. }The whole pipeline of our method includes three training phases. We first train a multi-step multi-view diffusion model on the Objaverse dataset, by fine-tuning an internal DiT~\cite{peebles2023scalable} based text-to-image model. We train this model for 30k iterations with 32 80G A100 GPUs using a total batch size of 128 and a learning rate of $3e^{-5}$. Then we perform distillation to distill the multi-step multi-view generator into a few-step multi-view generator, which takes 10k iterations with 32 80G A100 GPUs with a global batch size of 128 and a learning rate of $5e^{-6}$. Finally, we train a reconstruction model -- latent GS-LRM -- from scratch, which takes 80k iterations with 32 80G A100 GPUs with a total batch size of 256 and a learning rate of $4e^{-4}$. 

\subsection{Evaluation against baselines}
\label{main_res}

\begin{figure*}
    \centering
    \includegraphics[width=0.95\linewidth]{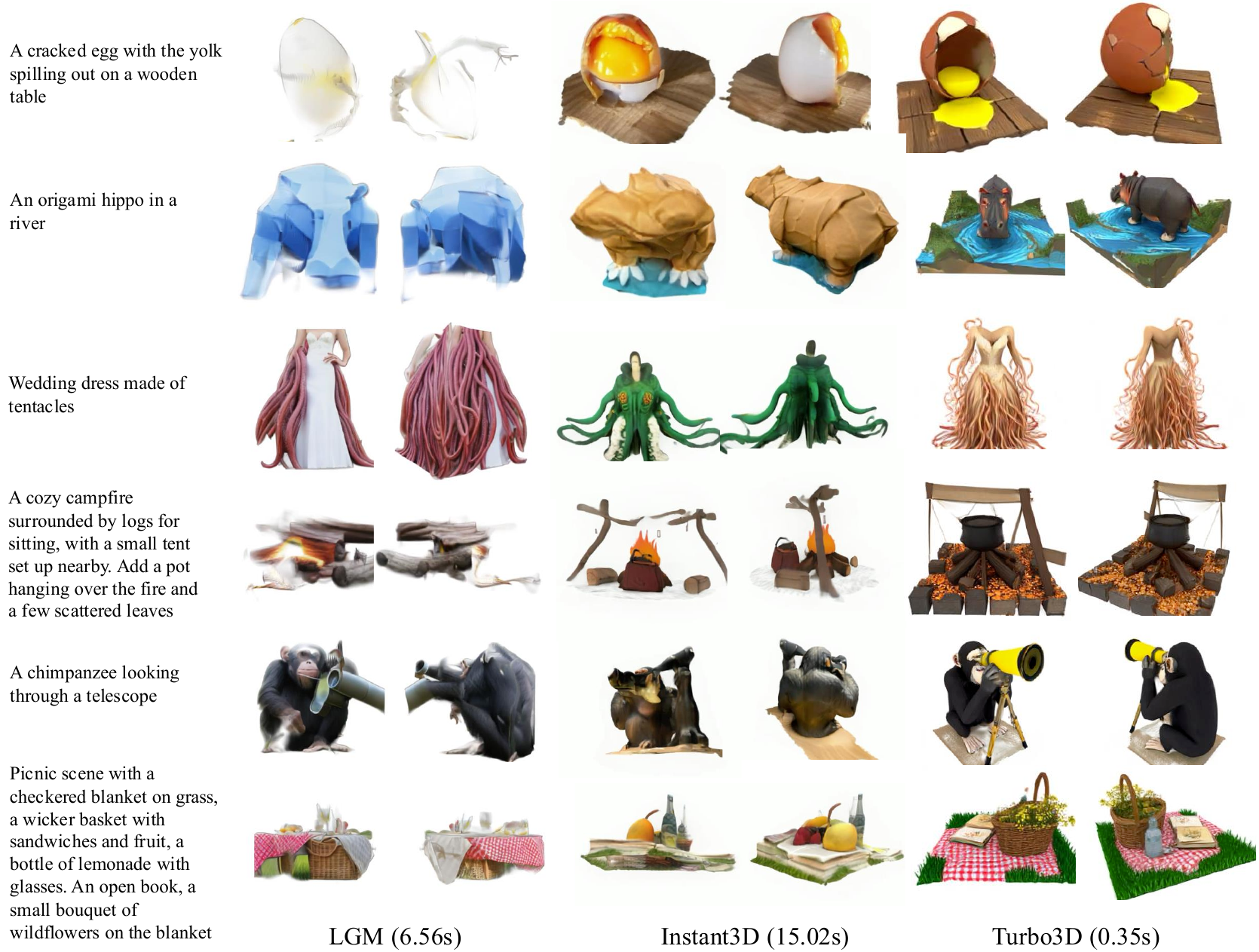}
    \caption{\textbf{Comparison of our \ours\ against baselines LGM~\cite{tang2025lgm} and Instant3D~\cite{li2023instant3d}.} Among these methods, Our method generates the most detailed and physically plausible 3D assets, closely adhering to the provided text prompts. In contrast, LGM tends to generate broken assets with Janus issue~\cite{poole2022dreamfusion}, while Instant3D has poorer text alignment, oftentimes missing some concepts, e.g., `spilling out' in the first row, `river' in the second row,  etc.   }
    \label{fig:comparison}
\end{figure*}

\begin{figure}
    \centering
    \includegraphics[width=1.0\linewidth]{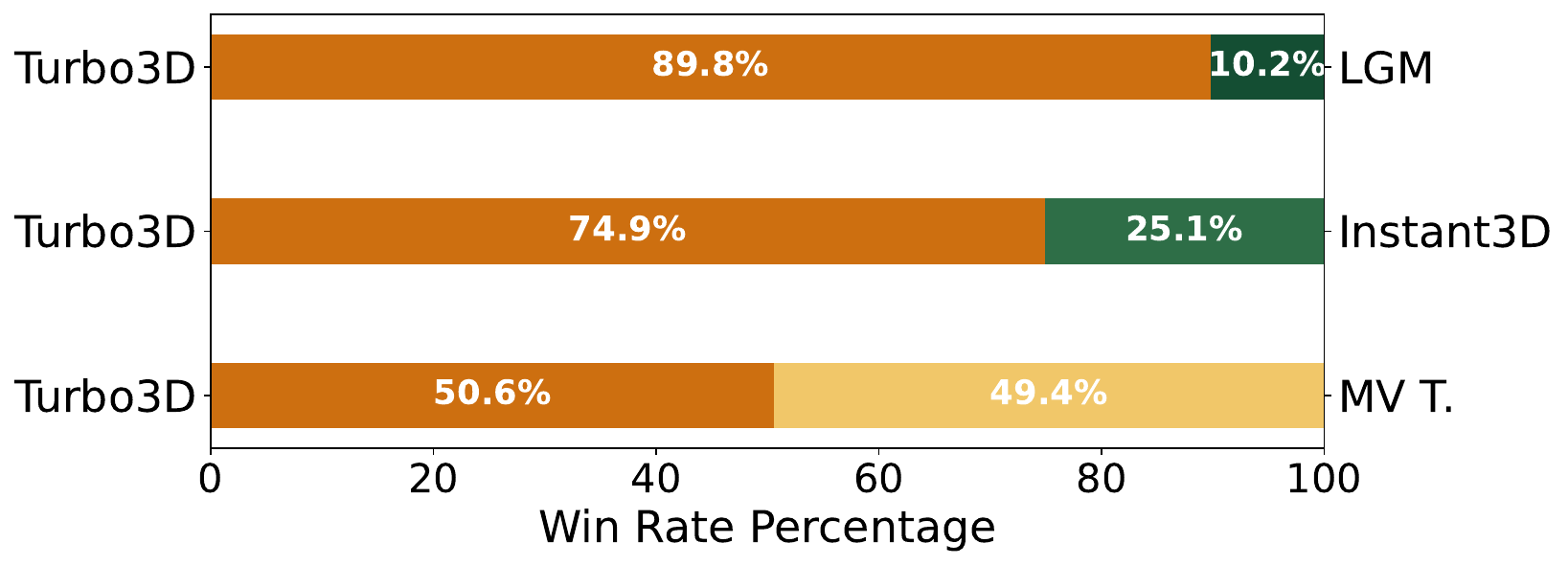}
    \caption{\textbf{User study results comparing our \ours\ to baseline LGM~\cite{tang2025lgm}, Instant3D~\cite{li2023instant3d}, and our slow MV teacher. } Our \ours\ is consistently preferred over baseline LGM and Instant3D, while having on-par preference with our MV teacher. See Fig.~\ref{fig:comparison},\ref{fig:ablation} for visual comparison.} 
    \label{user-study}
    \vspace{-2em}
\end{figure}

\begin{figure*}
    \centering
    \includegraphics[width=0.9\linewidth]{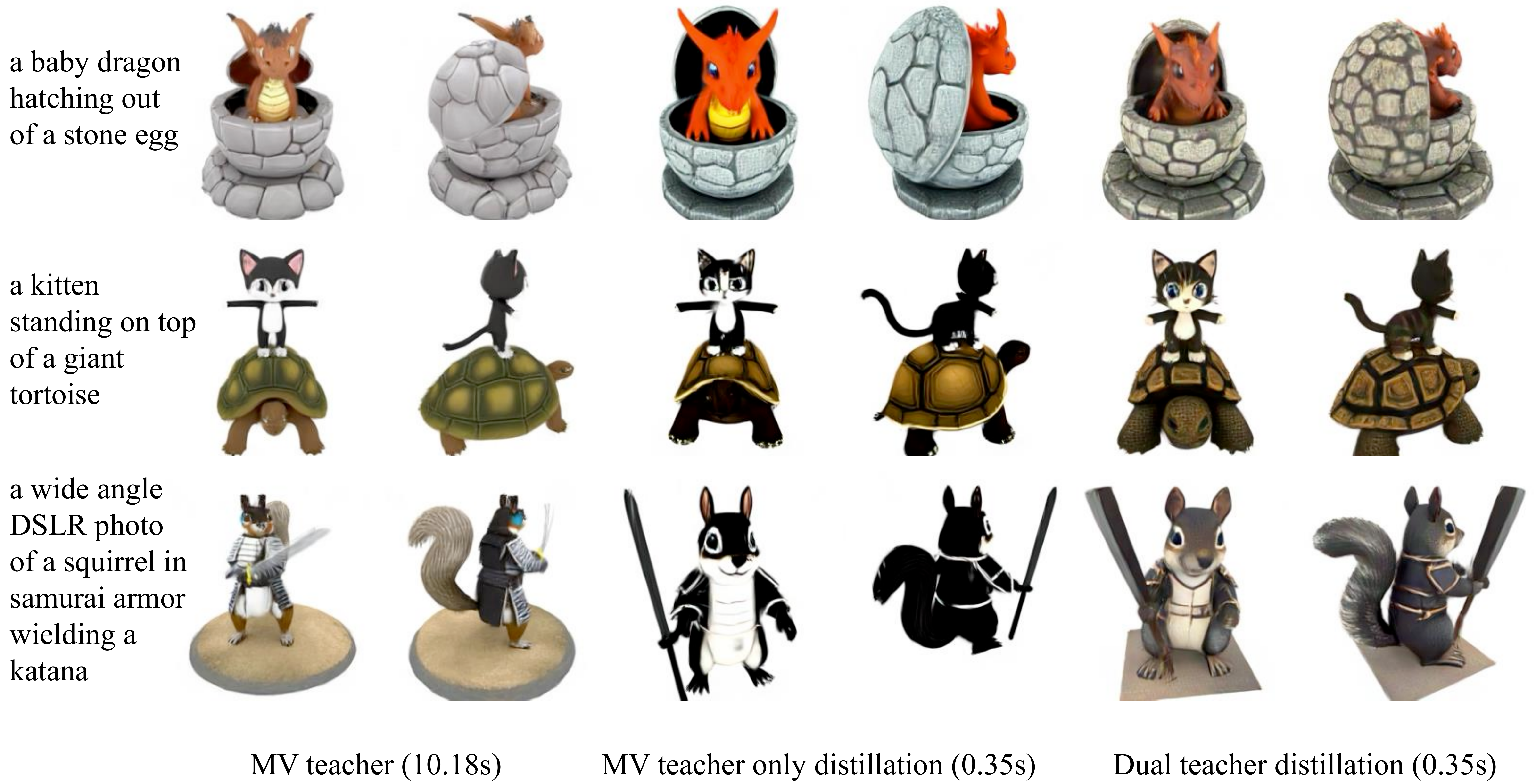}
    \caption{\textbf{Ablation of our Dual-teacher distillation algorithm.} Naively distilling MV teacher (middle column) causes compound mode collapse (see Sec.~\ref{sec:dual-teacher}), producing overly smooth synthetic-looking assets. Our dual-teacher distillation (right column) fixes the issue and generates 3D assets that are as photorealistic as, if not more than, the baseline MV teacher (left column). We also include the inference timings for each method; the distilled model is $\sim$50x faster than the teacher model. }
    \label{fig:ablation}
    \vspace{-1em}
\end{figure*}

\begin{table}[t]
\begin{center}
\begin{tabularx}{8.2cm}{p{2.5cm} X<{\centering} X<{\centering} X<{\centering}}
\toprule[1.5pt]
Method  & Clip Score $\uparrow$ & VQA Score $\uparrow$ & Inference Time $\downarrow$ \\
\midrule[1pt]
TripoSR~\cite{tochilkin2024triposr} & 23.85 & 0.57 & \second 1.19s\\
SV3D~\cite{voleti2024sv3d} & 24.92 & 0.64& 12.52s \\
\midrule
Instant3D~\cite{li2023instant3d} & \second 26.23 & \second 0.65& 15.02s\\
LGM~\cite{tang2025lgm} & 24.73 & 0.58 & 6.56s \\

Turbo3D (Ours)&  \first \textbf{27.61} & \first \textbf{0.76} & \first \textbf{0.35s}\\
\bottomrule[1.5pt]
\end{tabularx}
\end{center}
\vspace{-2mm}
\caption{\textbf{Comparison against state-of-the-art 3D generation methods.} Our \ours\ generates 3D assets with highest CLIP and VQA scores while using the least amount of time (benchmarked on a A100 GPU). }
\vspace{-2em}
\label{main}
\end{table}
\noindent\textbf{Qualitative Comparisons. } As shown in Fig.~\ref{fig:comparison}, our method generates significantly better results compared with LGM~\cite{tang2025lgm} and Instant3D~\cite{li2023instant3d}. In particular, LGM tends to produce simple 3D assets without geometry and texture details; it also lacks robustness, oftentimes generating low-quality 3D assets that are broken, or suffer from multi-face Janus problem~\cite{poole2022dreamfusion}, or do not closely follow provided text prompts. Instant3D is much more robust and able to produce plausible 3D assets most of the time. However, its performance on text-image coherence is limited compared with our \ours.
For example, in the first example, LGM’s rendition appears oversimplified with indistinct egg and yolk boundaries, while Instant3D provides a closer match but lacks fine details in the egg's structure and wood texture and fails to capture the concept of `spill out'. For complex prompts that describes a scene with multiple objects (4th and 6th rows), LGM generates a Janus asset while Instant3D miss a lot of concepts like blanket, basket, wildflowers etc; our generations adhere to the prompts much more closely.  These qualitative results highlight our \ours’s ability to generate high-quality, text-aligned 3D models with a superior level of detail, realism, and coherence across diverse and complex prompts, outperforming both LGM and Instant3D by a large margin.

\noindent\textbf{Quantitative Comparisons. }
Tab.~\ref{main} presents a quantitative comparison of our proposed method, Turbo3D, against several state-of-the-art approaches, including TripoSR~\cite{tochilkin2024triposr}, SV3D~\cite{voleti2024sv3d}, Instant3D~\cite{li2023instant3d}, and LGM~\cite{tang2025lgm}, on the text-to-3D task. 
Our proposed Turbo3D achieves the highest CLIP Score of 27.61 and VQA Score of 0.76, outperforming other methods by a significant margin in both quality metrics. In addition to the quality improvement, Turbo3D demonstrates remarkable efficiency with an inference time of only 0.35 seconds, substantially faster than competing methods. Although TripoSR is able to generate a 3D asset in only 1.19s, the quality of the generated results is highly limited. As a result, our proposed Turbo3D is able to achieve outstanding performance in terms of both quality and inference speed compared with state-of-the-art methods.

\noindent\textbf{User Study. }We run a user study by randomly selecting 80 text prompts and asking 56 users to make 1120 pairwise comparisons (users are shown an input text prompt and two generated 3D assets from two anonymous methods). We show results in Fig.~\ref{user-study}. Since the quantitative results of TripoSR is not competitive with others, it is not included in the user study. Moreover, SV3D only outputs videos instead of 3D representations of objects, so it is hard to perform a fair comparison. Therefore, we compare our Turbo3D with LGM~\cite{tang2025lgm}, Instant3D~\cite{li2023instant3d}, and our MV Teacher. 
In particular, MV Teacher is the model which our Turbo3D gets distilled from. When compared to LGM, Turbo3D achieved a win rate of 89.8\%, with only 10.2\% of participants favoring LGM. Against Instant3D, Turbo3D also outperformed with a 74.9\% win rate, indicating that users consistently found Turbo3D’s outputs more aligned with the input text descriptions and more visually compelling. When evaluated against our teacher model, Turbo3D held a close win rate of 50.6\%, with 49.4\% preferring MV T., reflecting comparable quality between the student and teacher models. The result indicates that our Turbo3D not only achieves significant speedup with distillation, but also preserves the generation ability from the teacher model.

\subsection{Ablation Study}
\label{ablation-study}

\noindent\textbf{Effect of Single-view Teacher. }
In Tab.~\ref{abla-dual}, we demonstrate the effectiveness of the dual teacher distillation strategy. The first line is our multi-view (MV) teacher model, which can achieve impressive results but runs  slowly because of the many diffusion sampling steps required. When performing distillation only with this MV Teacher, the quality drops by a large margin as shown in the second row. When adding a single-view teacher model for distillation, the distilled model is able to achieve much better results compared with the previous one. This configuration approaches the performance of the Multi-step MV Model while maintaining the efficiency benefits of the few-step setup, showcasing the advantages of using a dual-teacher strategy in our distillation. We also provide visualizations of the three models in Fig.~\ref{fig:ablation}. These comparisons demonstrate that the dual teacher distillation model strikes a balance between detail retention and stylization, closely replicating the quality of the MV teacher model while benefiting from the efficiency gains of distillation.

\noindent\textbf{Effect of Latent GS-LRM. }
In Tab.~\ref{abla-gslrm}, we showcase the effectiveness of latent GS-LRM. Compared with the original GS-LRM which operates in pixel space, our latent GS-LRM is able to skip the expensive image decoding process
while achieving similar image quality. 
\begin{table}[t]
\begin{center}
\begin{tabularx}{8.2cm}{p{4.5cm} X<{\centering} X<{\centering} }
\toprule[1.5pt]
Ablation  & CLIP Score $\uparrow$ & VQA Score $\uparrow$  \\
\midrule[1pt]
Multi-step MV Model & \first \textbf{28.04} & \first \textbf{0.77} \\
\midrule[1pt]
Few-step Model (MV Teacher) & 26.60 & 0.69\\

Few-step Model (Dual Teacher)&  \second 27.61 & \second 0.76 \\
\bottomrule[1.5pt]
\end{tabularx}
\end{center}
\caption{\textbf{Ablation of dual teacher distillation}. Distillation leads to quality drop compared with the MV teacher model. Compared with naively distilling the single MV teacher, dual-teacher distillation leads to much smaller quality drop. See Fig.~\ref{fig:ablation} for visual comparison.}
\label{abla-dual}
\end{table}

\begin{table}[t]
\begin{center}
\begin{tabularx}{8.2cm}{p{3cm} X<{\centering} X<{\centering} X<{\centering}}
\toprule[1.5pt]
Ablation  & CLIP Score $\uparrow$ & VQA Score $\uparrow$ & Inference Time $\downarrow$ \\
\midrule[1pt]
Pixel GS-LRM~\cite{zhang2025gs} & \first \textbf{27.62} & \first \textbf{0.76} & 0.45s\\
Latent GS-LRM & 27.61 & \first \textbf{0.76} & \first \textbf{0.35s} \\

\bottomrule[1.5pt]
\end{tabularx}
\end{center}
\caption{\textbf{Ablation for latent GS-LRM.} We report the CLIP score, VQA score, and overall text-to-3D inference time for comparison. Our latent GS-LRM achieves similar image quality while enabling better efficiency ($\sim$22\% speedup). }
\label{abla-gslrm}
\vspace{-1em}
\end{table}

\section{Conclusion}
In this work, we propose \ours\ for ultra-fast text-to-3D generation. To enable fast multi-view generation, we propose to distill a multi-step multi-view generator into a few-step multi-view generator. Moreover, to restore the multi-view consistency and photo-realism during distillation, we introduce a novel dual-teacher distillation framework. To further improve the multi-view reconstruction efficiency, we propose a latent GS-LRM which directly reconstructs 3D Gaussians from multi-view latents. Extensive experiments demonstrate that our proposed Turbo3D is able to achieve outstanding performance in terms of both generation quality and inference efficiency.
\section*{Acknowledgements}
This work began during Hanzhe Hu and Tianwei Yin's internships at Adobe Research. This research was supported in part by NSF award IIS-2345610.
{
    \small
    \bibliographystyle{ieeenat_fullname}
    \bibliography{main}
}
\clearpage
\maketitlesupplementary


\section{Details of Multi-step Multi-view Generation Model}
\label{mv}
We directly fine-tune an internal DiT~\cite{peebles2023scalable} based text-to-image model into a text-to-multiview model. We fine-tune the model on the Objaverse dataset~\cite{deitke2023objaverse}. For the generation task, we render the dataset at a fixed elevation (20 degrees) and 16 equidistant azimuths. We empirically find that training with random views performs better than fixed views. In particular, during training, we randomly sample $f$ views from the rendered 16 views for each instance, where $f$ can be 4 or 8. Each view is conditioned on the corresponding Plücker embedding. For inference, we only infer 4 views for efficiency. 

\section{Experiments on 512 resolution}
\label{512}
Some of the previous methods (Instant3D and SV3D) generate results with a higher resolution of 512. For a fair comparison,
we also perform experiments on 512 resolution. Tab.~\ref{main-512} presents the quantitative comparisons with several state-of-the-art methods, where inference time is all measured under the resolution of 512. We can see our Turbo3D-512 version performs slightly better than our Turbo3D (256 resolution) with longer inference time, while outperforming other state-of-the-art methods by a large margin in terms of Clip score, VQA score, and inference speed.  

Tab.~\ref{lrm-512} displays the effectiveness of latent GS-LRM. Under a higher resolution, the speed-up gain for latent GS-LRM gets larger. Overall, the latent GS-LRM archives a speed-up of 0.34s for the whole text-to-3D process.

\section{Details of User Study}
\label{user}
\begin{figure}
    \centering
    \includegraphics[width=0.56\linewidth]{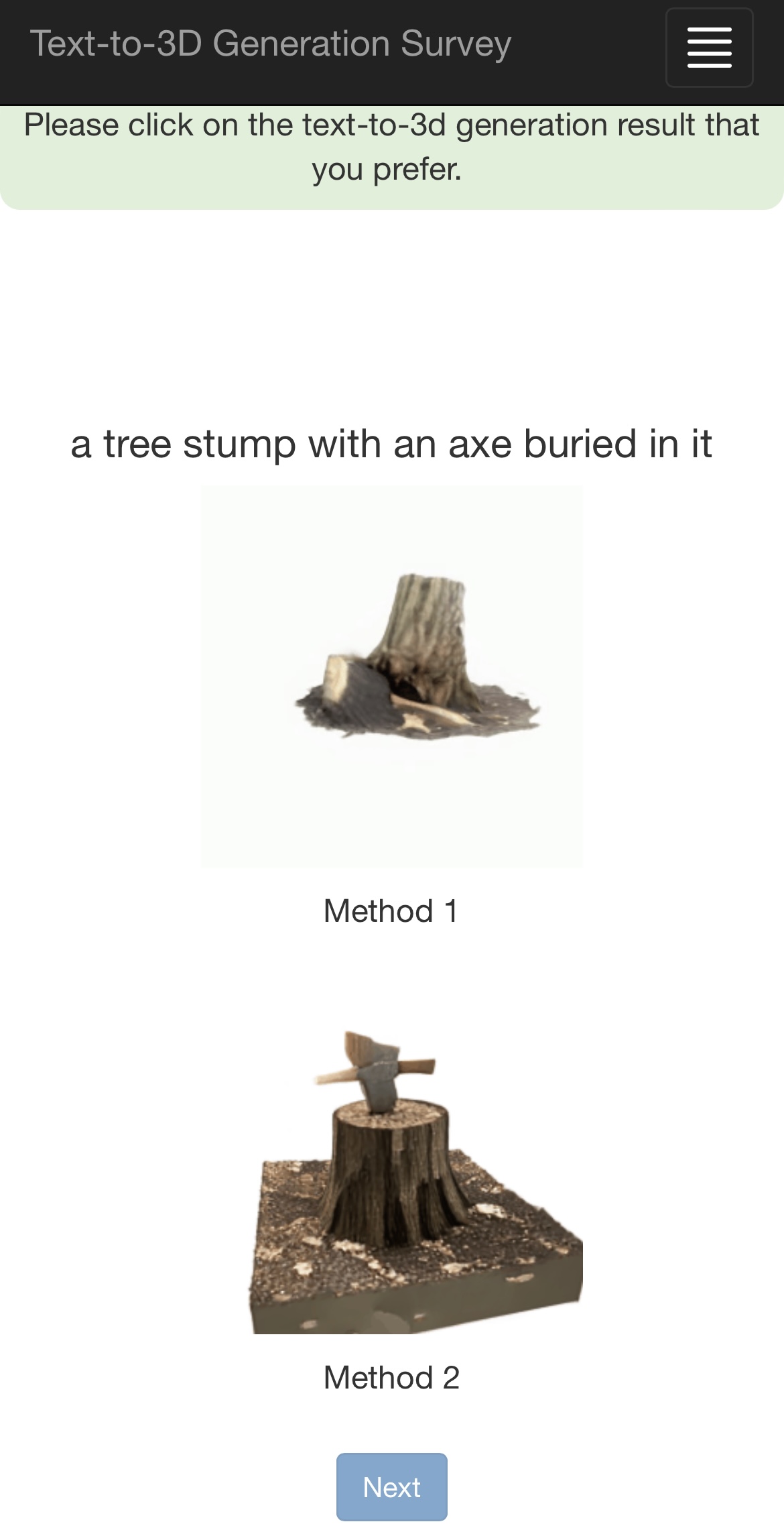}
    \caption{Interface example for user study.}
    \label{interface}
\end{figure}

The interface example is shown in Fig.~\ref{interface}. For each question, we show two rendered videos from two different methods and ask the user to pick their preferred one. The two methods are randomly chosen from the total 4 methods: LGM~\cite{tang2025lgm}, Instant3D~\cite{li2023instant3d}, our multi-step multi-view model and our Turbo3D.

\begin{table}[t]
\begin{center}
\begin{tabularx}{8.2cm}{p{2.5cm} X<{\centering} X<{\centering} X<{\centering}}
\toprule[1.5pt]
Method  & CLIP Score $\uparrow$ & VQA Score $\uparrow$ & Inference Time $\downarrow$ \\
\midrule[1pt]
TripoSR~\cite{tochilkin2024triposr} & 23.85 & 0.57 & \first 1.28s\\
SV3D~\cite{voleti2024sv3d} & 24.92 & 0.64& 35.96s \\
\midrule
Instant3D~\cite{li2023instant3d} & \second 26.23 & \second 0.65& 20.00s\\
LGM~\cite{tang2025lgm} & 24.73 & 0.58 & 6.56s \\

Turbo3D-512&  \first \textbf{27.66} & \first \textbf{0.78} & \first \textbf{1.28s}\\
\bottomrule[1.5pt]
\end{tabularx}
\end{center}
\caption{\textbf{Comparison against state-of-the-art 3D generation methods.} Our Turbo3D-512 generates 3D assets with highest CLIP and VQA scores while using the least amount of time (benchmarked on a A100 GPU). }
\label{main-512}
\end{table}

\begin{table}[t]
\begin{center}
\begin{tabularx}{8.2cm}{p{3cm} X<{\centering} X<{\centering} X<{\centering}}
\toprule[1.5pt]
Ablation  & CLIP Score $\uparrow$ & VQA Score $\uparrow$ & Inference Time $\downarrow$ \\
\midrule[1pt]
Pixel GS-LRM~\cite{zhang2025gs} & \first \textbf{27.68} & \first \textbf{0.78} & 1.62s\\
Latent GS-LRM & 27.66 & \first \textbf{0.78} & \first \textbf{1.28s} \\

\bottomrule[1.5pt]
\end{tabularx}
\end{center}
\caption{\textbf{Comparison between pixel and latent GS-LRM.} We report the CLIP score, VQA score, and overall text-to-3D inference time for comparison. Our latent GS-LRM achieves similar image quality while enabling better efficiency ($\sim$21\% speedup). }
\label{lrm-512}
\end{table}

\end{document}